\documentclass[11pt]{article}
\usepackage{acl2012,amsmath,caption,comment,graphicx,latexsym,multirow,spverbatim,subcaption,times}
\usepackage[hyphens]{url}

\title{SQUINKY! A Corpus of Sentence-level Formality,\\Informativeness, and Implicature}

\author{Shibamouli Lahiri \\
  Computer Science and Engineering \\
  University of Michigan \\
  Ann Arbor, MI 48109 \\
  {\tt lahiri@umich.edu}}

\date{}

\begin{document}

\maketitle

\begin{abstract}
We introduce a corpus of 7,032 sentences\footnote{A more recent (Aug 30, 2016) version of this paper appears at \url{http://web.eecs.umich.edu/~lahiri/new_draft.pdf}.} rated by human annotators for formality, informativeness, and implicature on a 1-7 scale. The corpus was annotated using Amazon Mechanical Turk.\footnote{\url{https://www.mturk.com/mturk/welcome}.} Reliability in the obtained judgments was examined by comparing mean ratings across two MTurk experiments, and correlation with pilot annotations (on sentence formality) conducted in a more controlled setting. Despite the subjectivity and inherent difficulty of the annotation task, correlations between mean ratings were quite encouraging, especially on formality and informativeness. We further explored correlation between the three linguistic variables, genre-wise variation of ratings and correlations within genres, compatibility with automatic stylistic scoring, and sentential make-up of a document in terms of style. To date, our corpus is the largest sentence-level annotated corpus released for formality, informativeness, and implicature.
\end{abstract}

\section{Introduction}
\label{sec:intro}

Consider the two following utterances:\footnote{Courtesy: \url{http://www.word-mart.com/html/formal_and_informal_writing.html}}
\begin{enumerate}
\item \begin{spverbatim}
This is to inform you that your book has been rejected by our publishing company as it was not up to the required standard. In case you would like us to reconsider it, we would suggest that you go over it and make some necessary changes.
\end{spverbatim}
\item \begin{spverbatim}
You know that book I wrote? Well, the publishing company rejected it. They thought it was awful. But hey, I did the best I could, and I think it was great. I'm not gonna redo it the way they said I should.
\end{spverbatim}
\end{enumerate}

Not only are the styles of the two utterances different (first one is formal, second one is informal), but they are also targeted at different people. This dichotomy of (in)formal expressions was examined in great detail by \newcite{Heylighen99formalityof}. As they observed, formality is the most important dimension of writing style (cf. \cite{variation_across_speech_and_writing,jstor_hudson_1994}),\footnote{For a general discussion on the theory of registers, see \cite{Levelt89} and \cite{leckie1995language}.} and has close connections to informativeness and implicature. They argued, in particular, that formality emerges out of a communicative objective -- 
to maximize the amount of information being conveyed to the listener while at the same time maintaining (or at least appearing to maintain) Grice's communicative maxims of Quality, Quantity, Relevance and Manner as much as possible \cite{grice1975logic}.

Heylighen and Dewaele introduced the notion of \emph{deep formality} -- ``avoidance of ambiguity by minimizing the context-dependence and fuzziness of expression'', and reasoned that the other type of formality (\emph{surface formality}; formalizing language for stylistic effects) is a corruption of the language's original deep purpose. Deep formality was characterized by a lack of \emph{contextuality}, evidenced in particular by decreased levels of \emph{deixis} and \emph{implicature} in linguistic realizations.

While several of the arguments Heylighen and Dewaele made are open to question
, an important take-home message from their theory is a so-called \emph{continuum of formality}, arising out of a process where a document (or a piece of text) can be ``formalized'' \emph{ad infinitum}, simply by adding more and more context. This precludes us from labeling a document or a sentence binarily as ``formal'' or ``informal''. We will instead follow the Likert scale approach \cite{likert:1932} to sentence formality annotation, shown to work well by \newcite{DBLP:journals/corr/abs-1109-0069}. 
In some sense, our work is similar to the Stanford politeness corpus \cite{conf/acl/Danescu-Niculescu-MizilSJLP13}; both corpora are at the sentence/utterance level, and both measure a pragmatic variable on an ordinal scale (formality vs politeness).

\section{Background and Related Work}
\label{sec:background}

\subsection{Formality}
\label{subsec:formality_background}

Heylighen and Dewaele's study, while seminal in the field of formality scoring, had its limitations. Although they stressed the relationship between contextuality (missing information) and implicature, it was never quantified. They also refrained from quantifying implicature itself 
-- to ``avoid all intricacies at the level of phonetics, syntax, semantics and pragmatics'', citing that the ``recognition of phonetic patterns, syntactical parsing, and even more semantic and pragmatic interpretation of natural language are still extremely difficult\ldots to perform automatically.'' Further, we suspect that the relation between deep formality and implicature might have been over-emphasized (cf. Section \ref{subsec:relationship_with_others}).

In the end, they quantified formality using deixis only (percentage difference between deictic and non-deictic parts-of-speech), which we will henceforth refer to as the ``F-score''.\footnote{Not to be confused with the harmonic mean of precision and recall.} F-score was used in genre analysis by \newcite{nowson2005wga}, and shown to be quite effective in discriminating between the 17 genres used in their study. Further, systematic variation in F-score was observed across gender and personality traits. \newcite{conf/icwsm/Teddiman09} noted in particular that F-score can successfully differentiate between genres, but it cannot explain why the genres are different. 
F-score was found to be the same for diary entries, and comments on those entries.\footnote{This could be due to linguistic style co-ordination \cite{Danescu-Niculescu-Mizil:thesis2012}.} In follow-up work, \newcite{FLAIRS135899} proposed a version of F-score (called ``CF-score'') based on Coh-Metrix \cite{coh_metrix} dimensions of narrativity, referential and deep cohesion, syntactic simplicity and word concreteness. CF-score was better able to discriminate between genres than 
F-score.

In a separate strand of work, \newcite{brooke-hirst:2014:Coling} identified formality as a continuous lexical attribute, and assigned a formality score to a word based on its co-occurrece frequency with a hand-picked seed set of formal and informal words, smoothed by latent semantic analysis \cite{Brooke:2010:AAL:1944566.1944577}. Formality of words was further shown to be correlated with other stylistic dimensions such as concreteness and subjectivity \cite{brooke-hirst:2013:IJCNLP}.

While all the above studies are very important, they looked at formality from document and word levels, not from the sentence level. \newcite{Sheikha12learningto} equated formality of a sentence with the formality of its corresponding document, and \newcite{brooke-hirst:2014:Coling} predicted formality of sentences using word-level features. \newcite{Peterson:2011:EFW:2021109.2021120} and \newcite{machili2014writing} looked into formality of emails at workplace, the former exploring the Enron corpus and how formality varies with social distance, relative power, and the weight of imposition, and the latter conducting similar analyses among workplace emails from Greek multinational companies.

As \newcite{Lahiri:2011:IJS:1964750.1964792} showed in their work, sentence formality is \emph{not} the same as document formality. While it is true that sentences do follow document-level trends
, it was observed that there is a wide spread among sentences in terms of formality -- not all sentences from a document are equally formal (cf. \cite{DBLP:journals/corr/abs-1109-0069}, and Section \ref{subsec:sentential_makeup} of this paper). \newcite{DBLP:journals/corr/abs-1109-0069} further showed that there are cases where the words in a sentence are formal, but the sentence as a whole is not (\emph{``For all the stars in the sky, I do not care.''}) -- thus raising questions regarding a straightforward application of lexical formality to explain sentence formality.\footnote{Also see the examples given by \newcite{Potts08HSK}.}

The only two studies we are aware of that 
looked into formality annotation of sentences
, are \cite{DBLP:journals/corr/abs-1109-0069}, and \cite{dethlefs-EtAl:2014:EACL}. Lahiri and Lu annotated 600 sentences by two undergraduate linguistics students on a Likert scale of 1-5. Inter-rater agreement was shown to improve substantially from binary annotations, which could be attributed to the \emph{continuum of formality} phenomenon described in Section \ref{sec:intro}. Dethlefs et al., on the other hand, were interested in formality from a natural language generation (NLG) perspective.\footnote{Note that the importance of formality in language generation has long been recognized \cite{Hovy:1990:PNL:80174.80176,abusheikha-inkpen:2011:ENLG}.} They annotated utterances using Amazon Mechanical Turk on three dimensions of style -- colloquialism (opposite of formality), politeness, and naturalness. A 1-5 Likert scale was used. The problem with this study is that the number of annotated sentences was quite limited, and they came from a restricted class of documents talking about restaurant reviews in a single city. This makes Dethlefs et al.'s corpus unsuitable for our purpose. We wanted a generic corpus of sentences annotated with formality ratings that could help build a sentence formality predictor, so we extended the work of \newcite{DBLP:journals/corr/abs-1109-0069} instead.

\subsection{Implicature}
\label{subsec:implicature_background}

A second issue with Heylighen and Dewaele's F-score is that it is unreliable on small documents, such as sentences and utterances (cf. \cite{Lahiri:2011:IJS:1964750.1964792}). It is therefore of interest to examine if the F-score correlates with human notion of formality at sentence level (cf. Section \ref{subsec:relationship_with_others}). But perhaps even more importantly, it shows a big limitation in the formulation of F-score: it is based on deixis only, and fails to take into account the \emph{amount of implicature} present in a sentence.

Note that in general, it is true that as we add more context to a document (or a sentence), it tends to become longer. The opposite is also true: as we rob a document (or sentence) of context, it tends to become shorter (\emph{contextual}). So it could be reasoned that sentences by themselves have a lot of un-stated context (as compared to a document), which are resolved by looking at neighboring sentences.\footnote{Much like resolving the meaning of a word by looking at neighboring words.} So if we could somehow estimate the amount of ``missing'' context in a sentence, we would be one more step ahead in assessing its true formality.

Quantifying the missing context is complicated by the fact that it depends on both deixis and implicature. While F-score gives a reasonable estimate of the amount of \emph{relative deixis} present in a sentence, it does not give any estimate of the amount of implicature. This forced us to rate sentences for the amount of implicature they carry (on Likert scale, because implicature is a continuous attribute \cite{degen2015}). This annotation process not only gave us implicature ratings, but also allowed us to look into how subjective the concept of implicature is (cf. Section \ref{subsec:annotation}).

Note that \newcite{degen2015} had already conducted a similar study on implicature annotation using Mechanical Turk. However, the focus of her study was on one particular type of implicature (\emph{some} but not \emph{all}), and the annotation process was not tied to formality or any other stylistic attribute. Also to be noted is the fact that our annotated corpus of 7,032 sentences is much larger than Degen's corpus of 1,363 utterances.

A general discussion of the vast literature on implicature (starting with \newcite{grice1975logic}, and expanded by \newcite{harnish:1976a1}, among others) is beyond the scope of this paper. Interested readers are referred to the excellent book by \newcite{Potts05BOOK} for a gentle introduction to the theory of \emph{conventional implicatures} (CIs), and to \cite{doi:10.1080/08351818609389263,benotti:tel-00541571,benotti_paper} for a discussion on \emph{causal implicatures}. Grice also introduced \emph{scalar implicatures} -- arguably the most prominent class of implicatures -- that equate ``some'' with ``not all'' for the sake of politeness. \newcite{Papafragou2003253} discussed the acquisition of scalar implicatures by children, and \newcite{carston:1998b} related scalar implicatures with relevance and informativeness -- a topic we will briefly visit in the next section.


Apart from \newcite{degen2015}, we are not aware of any work that specifically looked into implicature rating at sentence/utterance level. Degen's work, as we already pointed out, is not tied to formality scoring, so we used our own dataset of 7,032 sentences to rate for both formality and implicature.

\subsection{Informativeness}
\label{subsec:informativeness_background}

We also rated sentences for informativeness -- a trait \newcite{Heylighen99formalityof} identified with \emph{deep formality}, where language is formalized to communicate meaning more clearly and directly. We will test this hypothesis by checking if the formality of a sentence positively correlates with its informativeness (Section \ref{subsec:relationship_with_others}). Interestingly, \newcite{carston:1998b} independently arrived at a similar conclusion: ``informativeness principles\ldots give rise to\ldots a strengthening or narrowing down of the encoded meaning of the utterance.'' While Carston's specific argument was tied to scalar implicatures, it is not very far-fetched to see that the same argument would, in effect, also apply to \emph{deep formality} as evinced by Heylighen and Dewaele.

It is to be noted that the word \emph{informativeness} has different connotations in different settings. In the machine translation community, for example, the word \emph{informativeness} denotes a type of \emph{fidelity} measure to be applied to the translated text -- in order to verify how much content of the original text is preserved under the translation 
\cite{Rajman01automaticallypredicting}. Informativeness of \emph{words and phrases} is an important parameter in problems ranging from named entity detection \cite{Rennie:2005:UTI:1076034.1076095} to keyword extraction \cite{conf/ic3k/TimonenTTCH12}. Under this setting, informativeness is known as \emph{term informativeness} \cite{kireyev:2009:NAACLHLT09,wu-giles:2013:NAACL-HLT}. 
Interestingly, \newcite{Rennie:2005:UTI:1076034.1076095} pointed out that their term informativeness estimation approach would be especially helpful in ``extracting information from \emph{informal}, written communication'' (emphasis ours).

While all the above studies are important in their own right, and ground-breaking in some cases, we found none that specifically looked into informativeness rating of sentences in the context of formality, and there is no publicly available annotated dataset for \emph{sentence informativeness}. In this work, we will bridge the gap.

\section{Corpus Creation}
\label{sec:corpus_creation}

\subsection{Data}
\label{subsec:data}

Our data comes from the pioneering study of \newcite{Lahiri:2011:IJS:1964750.1964792}. They compiled four different datasets -- blog posts, news articles, academic papers, and online forum threads -- each consisting of 100 documents. For the blog dataset, they collected most recent posts from the top 100 blogs listed by Technorati\footnote{\url{http://technorati.com/}.} on October 31, 2009. For the news article dataset, they collected 100 news articles from 20 news sites (five from each). The articles were mostly from ``Breaking News'', ``Recent News'', and ``Local News'' categories, with no specific preference attached to any particular category.\footnote{The news sites were CNN, CBS News, ABC News, Reuters, BBC News Online, New York Times, Los Angeles Times, The Guardian (U.K.), Voice of America, Boston Globe, Chicago Tribune, San Francisco Chronicle, Times Online (U.K.), news.com.au, Xinhua, The Times of India, Seattle Post Intelligencer, Daily Mail, and Bloomberg L.P.} For the academic paper dataset, they randomly sampled 100 papers from the CiteSeerX\footnote{\url{http://citeseerx.ist.psu.edu/}.} digital library. For the online forum dataset, they sampled 50 random documents crawled from the Ubuntu Forums,\footnote{\url{http://ubuntuforums.org/}.} and 50 random documents crawled from the TripAdvisor New York forum.\footnote{\url{http://www.tripadvisor.com/ShowForum-g60763-i5-New_York_City_New_York.html}.} The blog, news, paper, and forum datasets had 2110, 3009, 161406 and 2569 sentences respectively.

We manually cleaned and sentence-segmented the blog, news, and forum datasets to come up with 7,032 unique sentences. The much larger and more complex \emph{paper dataset} was discarded, because manual cleansing and sentence segmentation of text data extracted from PDF was prohibitively time-consuming, and often unsuccessful because of spurious characters, words, and corrupted/missing segments of text.\footnote{Note that this manual cleaning was necessary for our annotation process, because we cannot expect our annotators to deal with corrupt/incomplete/inaccurate sentences.}

\subsection{Annotation}
\label{subsec:annotation}

\begin{table*}
\begin{center}
\begin{tabular}{|l|cccc|}
\hline
& \textbf{Overall} & \textbf{Blog} & \textbf{News} & \textbf{Forum} \\
\hline
Formality & 0.68 & 0.60 & 0.35 & 0.48 \\
Informativeness & 0.64 & 0.63 & 0.42 & 0.63 \\
Implicature & 0.14 & 0.19 & 0.09 & 0.11 \\
\hline
\end{tabular}
\end{center}
\caption{\label{tab:correlation_exp} Spearman's $\rho$ between the mean ratings obtained from our Mechanical Turk experiments. All results are statistically significantly different from zero, with p-value $<$ 0.0001.}
\end{table*}

\begin{table*}
\begin{center}
\begin{tabular}{|l|cccc|}
\hline
& \textbf{Overall} & \textbf{Blog} & \textbf{News} & \textbf{Forum} \\
\hline
MTurk Experiment 1 & 0.78 & 0.73 & 0.32* & 0.49 \\
MTurk Experiment 2 & 0.73 & 0.61 & 0.30* & 0.53 \\
\hline
\end{tabular}
\end{center}
\caption{\label{tab:old_data} Spearman's $\rho$ between the mean formality ratings from Mechanical Turk, and mean formality ratings from \newcite{DBLP:journals/corr/abs-1109-0069}. All results are statistically significantly different from zero, with p-value $<$ 0.0001. For the results marked with a *, their p-values are $<$ 0.01.}
\end{table*}

With the 7,032 sentences, we conducted two Mechanical Turk annotation experiments. In our first experiment, Turkers were requested to rate sentences on a 1-7 scale for formality, informativeness, and implicature. Each sentence was a HIT (Human Intelligence Task), and we requested five \emph{assignments} per HIT so that we could get five independent ratings for each sentence. We requested Turkers with English as first language in our HIT title\footnote{\tt{How formal is this sentence? English as first language required.}} and description,\footnote{\tt{This is a formality survey HIT, where we have three stylistic questions on an English sentence. Please do not enter if you do not have English as first language.}} but there was no easy way to ensure that it was indeed the case. As a quick fix, we required ``Turkers from US'' as qualification, and hoped that the average across five independent ratings will paint a better picture than any individual rating alone. Our instructions were minimal -- we started with the two examples given at the beginning of Section \ref{sec:intro} to prime the Turkers with the notion of \emph{formality}, and gave them a few more links to explore the concept on their own.\footnote{\url{http://www.engvid.com/english-resource/formal-informal-english/}, \url{http://dictionary.cambridge.org/us/grammar/british-grammar/formal-and-informal-language}, \url{http://www.englishspark.com/informal-language/}, \url{http://www.antimoon.com/how/formal-informal-english.htm}.} Then we told them to rate sentences on how formal they are. Turkers were requested to be \emph{consistent} in their ratings across sentences, and rate sentences independently of each other. The order of presentation of the sentences was scrambled so as to remove any potential sequence effect. In total, 527 Turkers participated in our first experiment.

\begin{table*}
\scriptsize
\begin{center}
\begin{tabular}{|l|l|l|}
\hline
& \textbf{High} & \textbf{Low} \\
& & \\
\hline
Formality & And in its middle-class neighborhoods, Baghdad is a city & Thanx! \\
& of surprising topiary sculptures: leafy ficus trees are & \\
& carved in geometric spirals, balls, arches and squares, as & \\
& if to impose order on a chaotic sprawl. & \\
& & \\
\hline
Informativeness & According to the Shanghai Jiao Tong University Press, & Any recommendations? \\
& the press is currently compiling a picture album of Qian & \\
& and a collection of his writings based on 800-plus-page & \\
& documents retrieved from the U.S. National Archives, & \\
& which include details about his encounters with the U.S. & \\
& government and his trip back home. & \\
& & \\
\hline
Implicature & Who will join? & Most mornings they rise before their rooster crows, bolting \\
& & down a meager breakfast of coconut and chile-spiced \\
& & vegetables over rice before venturing out on their journey: \\
& & rowing to school aboard a hand-carved 15-foot sampan. \\
& & \\
\hline
\end{tabular}
\end{center}
\caption{\label{tab:fii} Example sentences with high and low mean MTurk ratings for formality, informativeness, and implicature.}
\end{table*}

Note, however, that assessing inter-rater agreement becomes difficult on Mechanical Turk because different Turkers work on different number of HITs. Furthermore, we had no quality control other than ``US-based'' in our first experiment. This is why we conducted a second experiment, which was essentially identical to the first, except that now we added two more requirements -- at least 1,000 HITs completed with at least 99\% approval rate -- on top of the US-based requirement. This resulted in 187 Turkers participating in our second experiment.

Correlations between the mean ratings obtained from these two experiments are shown in Table \ref{tab:correlation_exp}. Several things are to be noted from this table. First, note that even without quality control (and weak enforcement of the English-first-language policy), Turkers' \emph{mean ratings} correlated pretty well (across two experiments) for both formality as well as informativeness, echoing previous findings by \newcite{DBLP:journals/corr/abs-1109-0069}. Second, it shows that even without extensive and detailed instructions, Turkers were able to rate subjective concepts like ``formality'' and ``informativeness'' quite well, again echoing the findings summarized by Lahiri and Lu. Note that we did not provide Turkers with extensive and detailed instructions because:

\begin{itemize}
\item We did not want to bias them with our view of the English language (removing \emph{experimenter bias}).
\item We wanted to see if Likert scale annotations were good enough (as claimed by \newcite{DBLP:journals/corr/abs-1109-0069}) to instil sufficient reliability and agreement in the annotation process, especially between mean ratings.
\item We wanted to see if mean ratings across multiple raters could effectively eliminate the idiosyncrasies of individual Turkers in a subjective annotation task like this.\footnote{Here are the three questions we asked: {\tt How formal do you think is the above sentence? How much information do you think the above sentence carries? How much do you think the above sentence implies/suggests, or leaves to possible interpretations?} We also had optional comment boxes so that Turkers can leave us their thoughts on the annotation process.}
\end{itemize}

\begin{figure*}
    \centering
    \begin{subfigure}{0.32\textwidth}
        \centering
        \includegraphics[width=\textwidth]{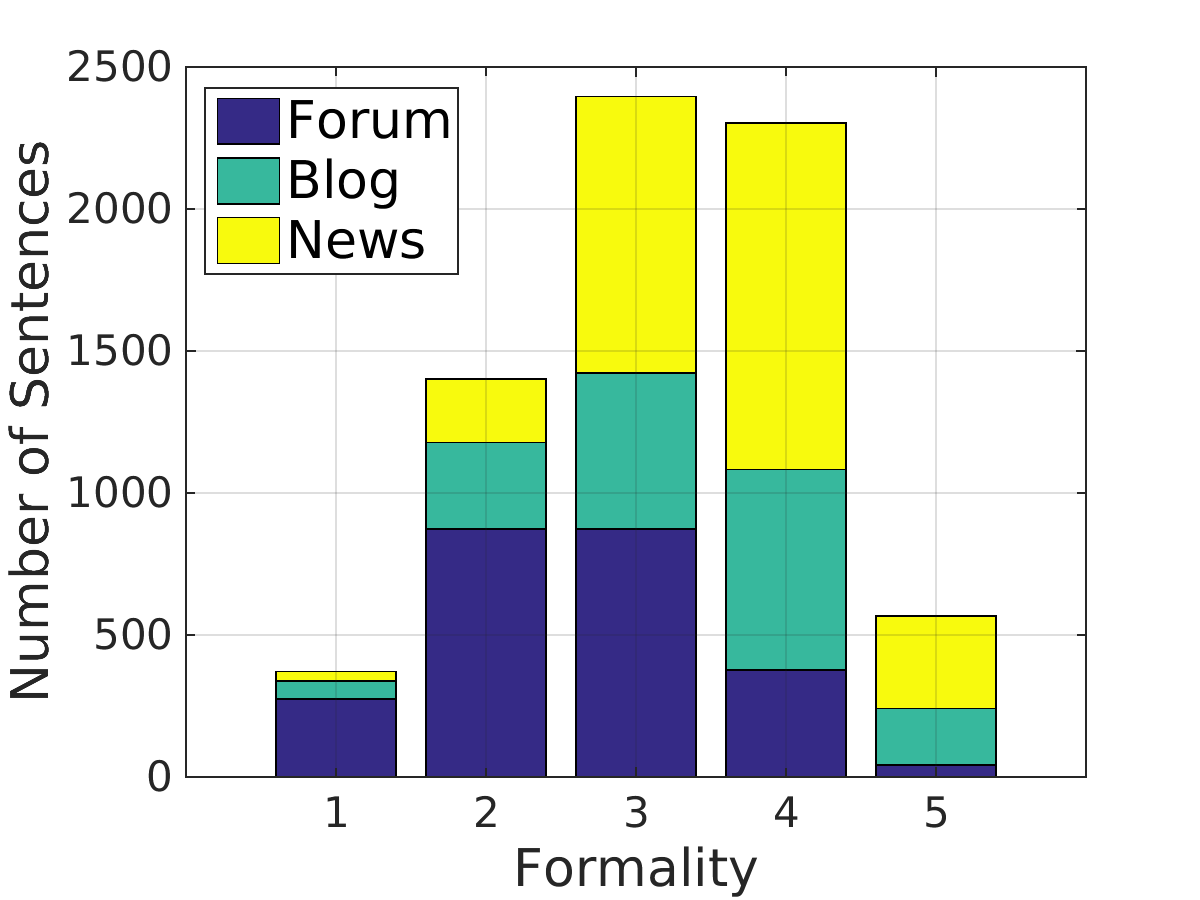}
        \label{fig:formality_genre}
    \end{subfigure}
\hfill
    \begin{subfigure}{0.32\textwidth}
        \centering
        \includegraphics[width=\textwidth]{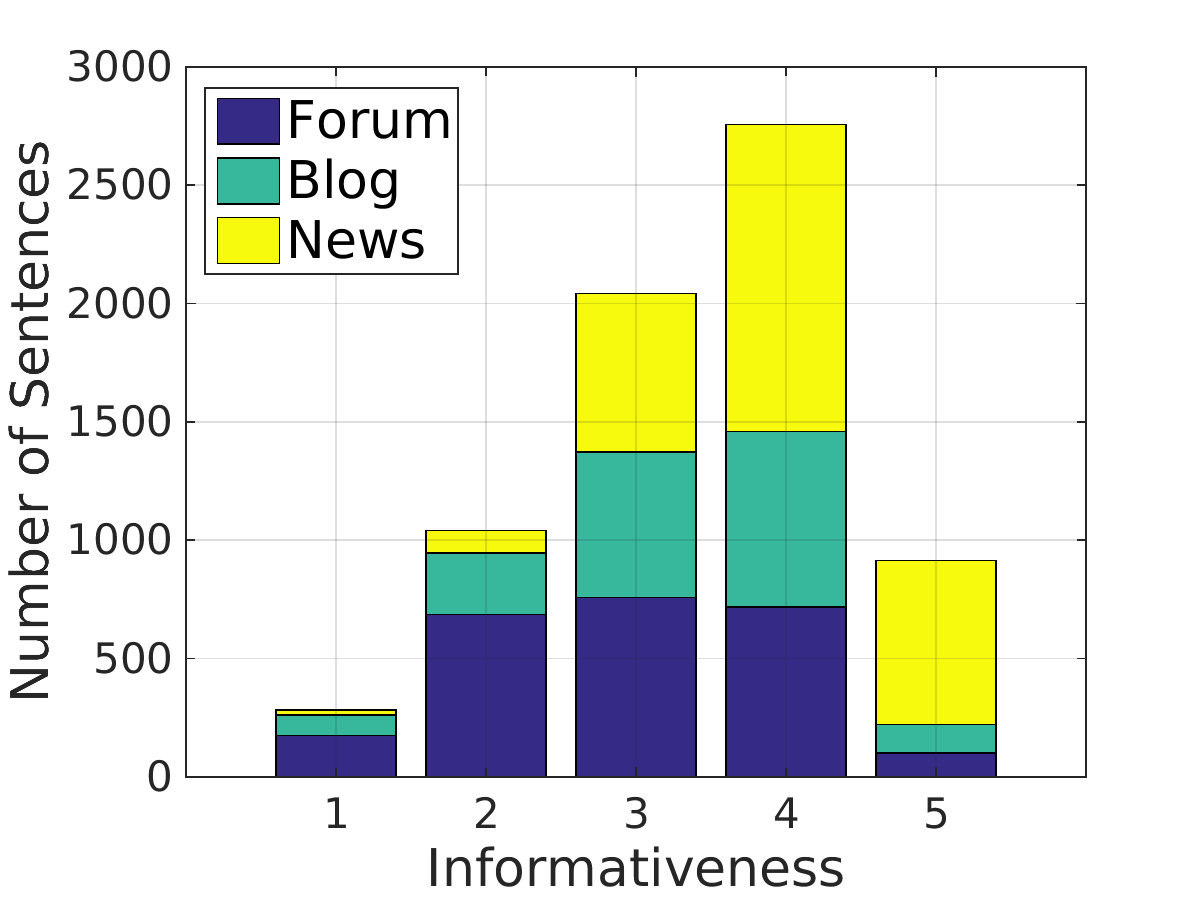}
        \label{fig:informativeness_genre}
    \end{subfigure}
\hfill
    \begin{subfigure}{0.32\textwidth}
        \centering
        \includegraphics[width=\textwidth]{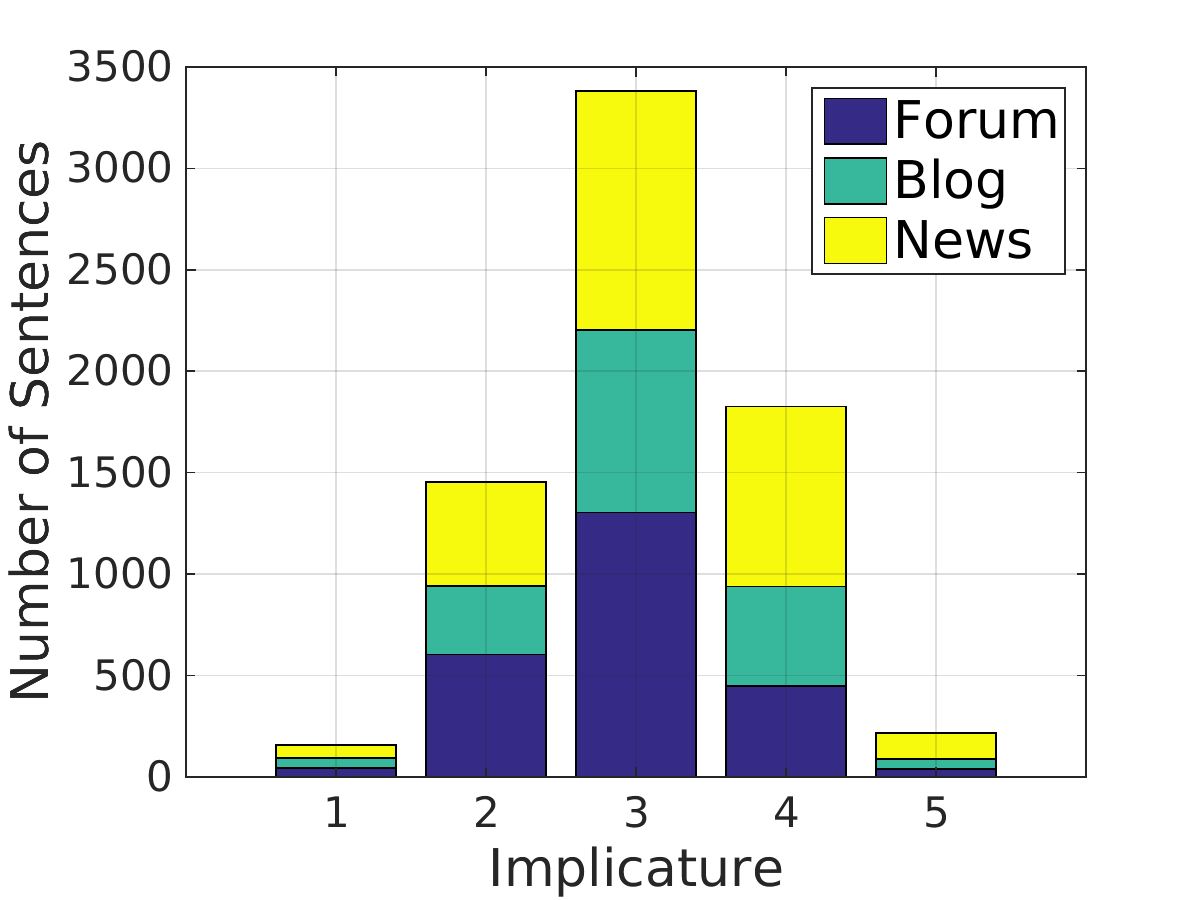}
        \label{fig:implicature_genre}
    \end{subfigure}
    \caption{Genre-wise variation of formality, informativeness, and implicature (can be viewed in grayscale).}
    \label{fig:fii_genre}
\end{figure*}

Having said that, note from Table \ref{tab:correlation_exp} that the correlation values for implicature are rather low -- across all genres (albeit positive). This is unsurprising, however, given that implicature is arguably the most subjective among the three pragmatic variables we investigated, and quite possibly, the least amenable to any straightforward syntactic, lexical, or semantic explanation.

We further compared our mean \emph{formality ratings} from Mechanical Turk to the mean formality ratings reported by \newcite{DBLP:journals/corr/abs-1109-0069} in their ``actual'' annotation phase. Results are shown in Table \ref{tab:old_data}. Note that the mean Turker ratings are highly positively correlated with the mean ratings from Lahiri and Lu's quality-controlled study -- except the \emph{news} genre, where correlations are weaker (also see Table \ref{tab:correlation_exp}). We plan to investigate the news genre in future work. But the overall patterns are strongly encouraging, and validate the idea that a formality-annotated corpus can indeed be built reliably with Likert-scale-style annotations.

We show some example high- and low- formality, informativeness and implicature sentences in Table \ref{tab:fii}.\footnote{The full dataset is available at \url{https://drive.google.com/file/d/0B2Mzhc7popBgdXZmRlg2RUdqdDA/view?usp=sharing}. Examples in Table \ref{tab:fii} are from our second MTurk experiment, which comprises better-qualified Turkers.} Note that they follow the usual intuitions about formality, informativeness, and implicature quite well; for example, sentences that are high in formality and informativeness, but low in implicature, are longer and more difficult to read. The opposite is also true; informal and uninformative sentences are much shorter, and are often laden with a lot of implicature.\footnote{Interesting trivia: the title of this paper derives from a sentence in our corpus that is very low in formality and informativeness, and medium in implicature.} For the rest of the paper, we only consider the mean ratings from our \emph{second MTurk experiment}, which comprises better-qualified Turkers. For notational convenience, \emph{mean ratings} will henceforth be referred to as \emph{Formality}, \emph{Informativeness} and \emph{Implicature}, as appropriate.

\section{Experiments}
\label{sec:experiments}

We performed three separate experiments on the 7,032 annotated sentences to identify different aspects of the annotations. In our first experiment, we explored how sentence-level formality, implicature, and informativeness vary across three different online genres -- news, blog, and forums (Section \ref{subsec:genre_wise_variation}). In the second experiment, we investigated the correlation among these three variables, and correlation with stylistic scores (Section \ref{subsec:relationship_with_others}). Finally, in Section \ref{subsec:sentential_makeup}, we examined how \emph{documents} varied in terms of sentential formality, informativeness, and implicature -- on average.

\begin{table*}
\scriptsize
\begin{center}
\begin{tabular}{lccccccccccccccccc}
& \multicolumn{8}{c}{\textbf{Overall}} && \multicolumn{8}{c}{\textbf{Blog}} \\
\hline
& \textbf{Fo} & \textbf{In} & \textbf{Im} & \textbf{Lw} & \textbf{Lc} & \textbf{F} & \textbf{I} & \textbf{LD} && \textbf{Fo} & \textbf{In} & \textbf{Im} & \textbf{Lw} & \textbf{Lc} & \textbf{F} & \textbf{I} & \textbf{LD} \\
\hline
\textbf{Fo} & 1.00 & 0.73 & 0.07 & 0.55 & 0.59 & 0.34 & 0.03* & \emph{0.01} && 1.00 & 0.73 & -0.10 & 0.51 & 0.54 & 0.33 & 0.07* & \emph{-0.04}  \\
\textbf{In} &  & 1.00 & 0.05 & 0.62 & 0.65 & 0.31 & 0.05 & \emph{-0.02}  &  && 1.00 & -0.08* & 0.62 & 0.65 & 0.29 & 0.06** & -0.06* \\
\textbf{Im} &  &  & 1.00 & 0.10 & 0.10 & -0.06 & 0.03** & \emph{0.00}  &  &  && 1.00 & \emph{0.02} & \emph{0.01} & -0.18 & \emph{0.04} & \emph{-0.02} \\
\textbf{Lw} &  &  &  & 1.00 & 0.98 & 0.23 & 0.12 & -0.18 &  &  &  && 1.00 & 0.98 & 0.18 & 0.13 & -0.23 \\
\textbf{Lc} &  &  &  &  & 1.00 & 0.28 & 0.07 & -0.08 &  &  &  &  && 1.00 & 0.23 & 0.08* & -0.15 \\
\textbf{F}  &  &  &  &  &  & 1.00 & -0.14 & 0.04*  &  &  &  &  &  && 1.00 & -0.12 & 0.06* \\
\textbf{I}  &  &  &  &  &  &  & 1.00 & \emph{-0.02}  &  &  &  &  &  &&  & 1.00 & -0.06** \\
\textbf{LD} &  &  &  &  &  &  &  & 1.00  &  &  &  &  &  &  &  && 1.00 \\
\hline
\multicolumn{18}{c}{}\\
& \multicolumn{8}{c}{\textbf{News}} && \multicolumn{8}{c}{\textbf{Forum}} \\
\hline
& \textbf{Fo} & \textbf{In} & \textbf{Im} & \textbf{Lw} & \textbf{Lc} & \textbf{F} & \textbf{I} & \textbf{LD} && \textbf{Fo} & \textbf{In} & \textbf{Im} & \textbf{Lw} & \textbf{Lc} & \textbf{F} & \textbf{I} & \textbf{LD} \\
\hline
\textbf{Fo} & 1.00 & 0.63 & -0.08 & 0.34 & 0.38 & 0.27 & \emph{-0.01} & \emph{0.00} && 1.00 & 0.57 & \emph{0.04} & 0.42 & 0.43 & 0.07* & 0.16 & -0.07* \\
\textbf{In} &  & 1.00 & -0.10 & 0.43 & 0.45 & 0.28 & \emph{-0.01} & \emph{-0.02}  &  && 1.00 & 0.08 & 0.58 & 0.60 & 0.09 & 0.16 & -0.08* \\
\textbf{Im} &  &  & 1.00 & \emph{-0.01} & \emph{-0.02} & -0.12 & \emph{0.02} & \emph{0.00}  &&  &  & 1.00 & 0.06* & 0.05* & -0.08* & 0.05** & \emph{-0.02} \\
\textbf{Lw} &  &  &  & 1.00 & 0.98 & 0.21 & 0.08 & -0.17 &  &  &  && 1.00 & 0.97 & \emph{0.02} & 0.23 & -0.26 \\
\textbf{Lc} &  &  &  &  & 1.00 & 0.27 & \emph{0.03} & -0.08 &  &  &&  &  & 1.00 & 0.06* & 0.19 & -0.15 \\
\textbf{F} &  &  &  &  &  & 1.00 & -0.15 & \emph{-0.03}  &  &  &  &&  &  & 1.00 & -0.12 & \emph{0.01} \\
\textbf{I} &  &  &  &  &  &  & 1.00 & 0.05*  &  &  &  &  &  &  && 1.00 & \emph{-0.03} \\
\textbf{LD} &  &  &  &  &  &  &  & 1.00  &  &  &  &  &  &  &  && 1.00 \\
\hline
\end{tabular}
\end{center}
\caption{\label{tab:big_corr} Spearman's $\rho$ between stylistic variables, as explained in text. Most of the results are statistically significantly different from zero, with p-value $<$ 0.0001. For the results marked with a *, p-values are $<$ 0.01; for those marked with a **, p-values are $<$ 0.05. Results in \emph{italics} are statistically insignificant.}
\end{table*}

\begin{figure*}
    \centering
    \begin{subfigure}{0.47\textwidth}
        \centering
        \includegraphics[width=\textwidth]{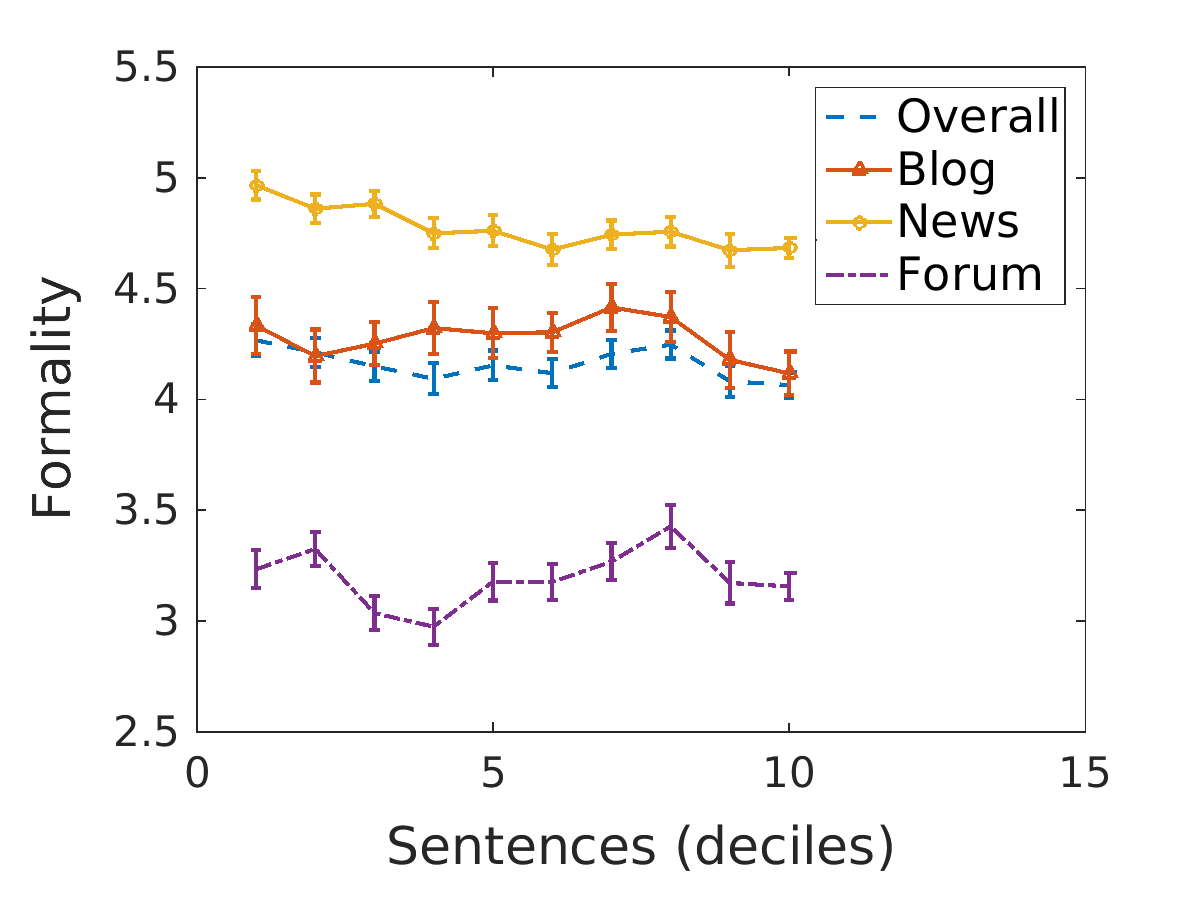}
        \label{fig:formality_sent}
    \end{subfigure}
\hfill
    \begin{subfigure}{0.47\textwidth}
        \centering
        \includegraphics[width=\textwidth]{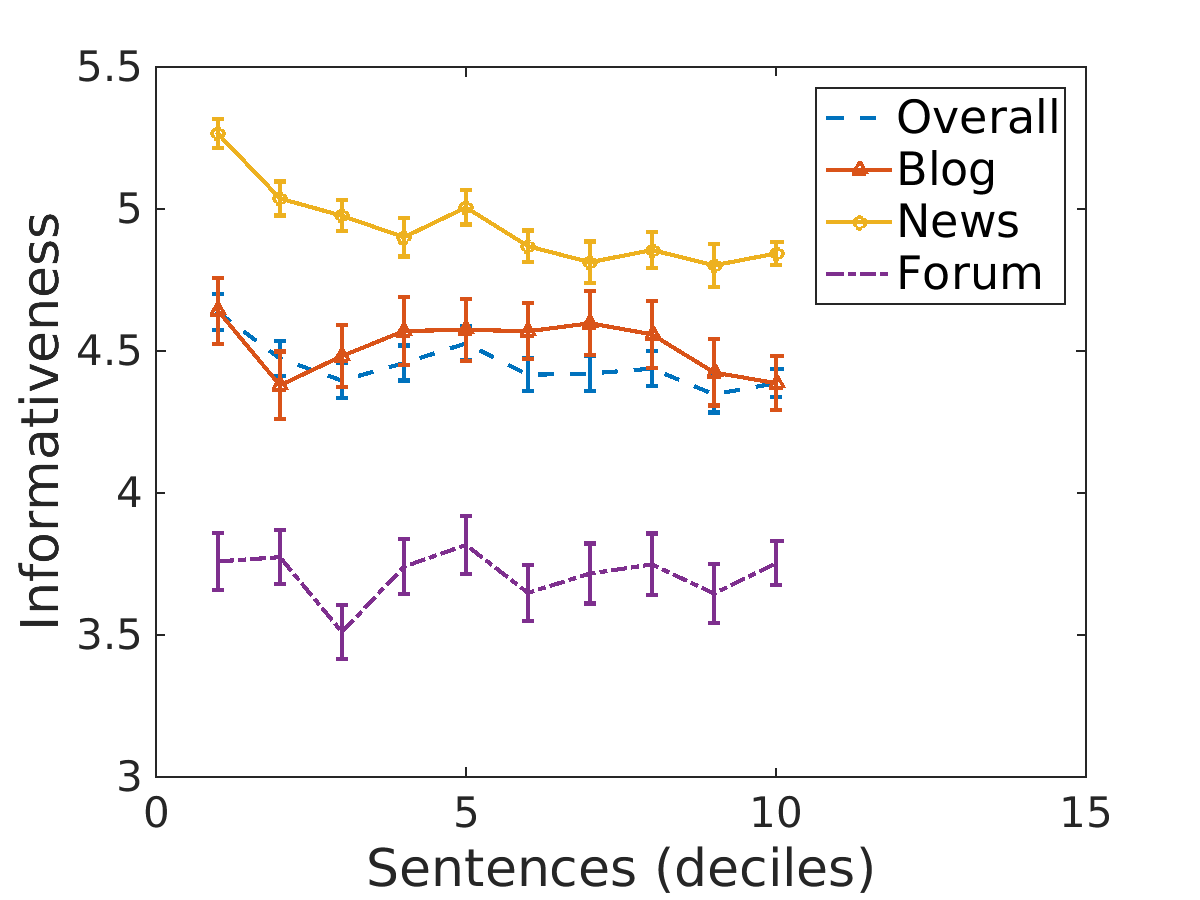}
        \label{fig:informativeness_sent}
    \end{subfigure}

    \begin{subfigure}{0.47\textwidth}
        \centering
        \includegraphics[width=\textwidth]{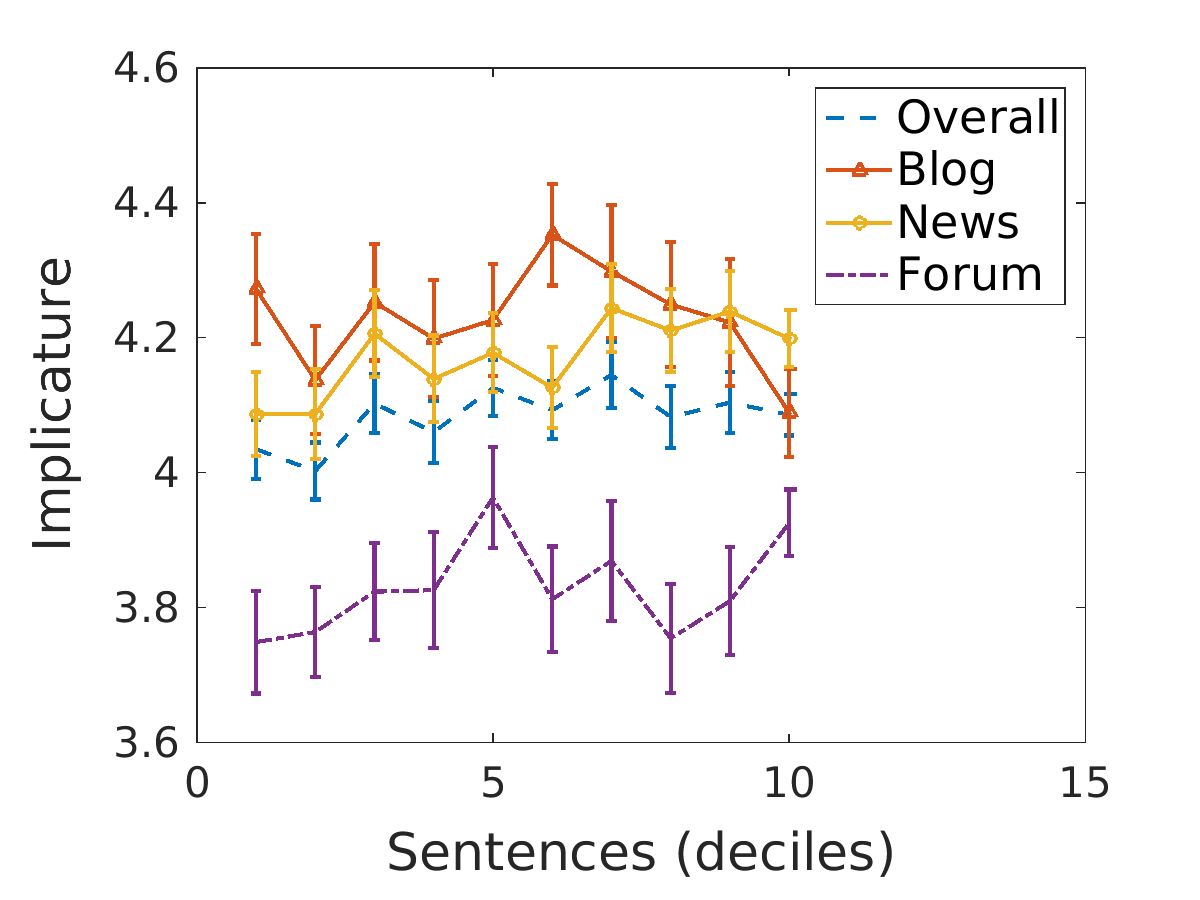}
        \label{fig:implicature_sent}
    \end{subfigure}
    \caption{Sentential make-up of formality, informativeness, and implicature (can be viewed in grayscale).}
    \label{fig:fii_sent}
\end{figure*}

\subsection{Genre-wise Variation}
\label{subsec:genre_wise_variation}

We plot five-bin histograms of formality, informativeness, and implicature in Figure \ref{fig:fii_genre}. Note from Figure \ref{fig:fii_genre} that \emph{overall}, our corpus is dominated by high-informativeness, mid-to-high-formality, and mid-implicature sentences. Since our implicature rating is less reliable than the other two ratings (cf. Section \ref{subsec:annotation}), it is relatively unclear whether this \emph{mid-implicature} trend is a real phenomenon, or is more of a reflection of \emph{central tendency bias} among the annotators -- who, lacking a better choice and a better interpretation -- chose middling values for the implicature rating. Central tendency in implicature is also observed for the three individual genres -- news, blog, forums.

The news genre is dominated by high-informativeness, and mid-to-high-formality sentences; blogs, too, are mostly high-formality and mid-to-high-informativeness sentences; on the other hand, forums are dominated by mid-to-low-formality sentences, and are spread out almost evenly when it comes to informativeness. The general trends corroborate earlier studies \cite{Lahiri:2011:IJS:1964750.1964792,DBLP:journals/corr/abs-1109-0069}.

The fact that forums are spread out in terms of (sentential) informativeness shows that there are all kinds of sentences in forums -- some are very informative, some are somewhat informative, and some are uninformative (e.g., help-eliciting setences such as ``help please!'', sentences expressing gratitude such as ``Thanks everybody!'', and suggestive sentences such as ``give it a shot.''). Filtering forum sentences by informativeness may be a useful first step towards effective mining of forum data.

\subsection{Relationship with Others}
\label{subsec:relationship_with_others}

We experimented with eight different sentential stylistic variables, as detailed below:

\begin{enumerate}
\item \textbf{Fo:} Formality of the sentence, i.e., the mean formality rating assigned by Turkers in our second MTurk experiment.
\item \textbf{In:} Informativeness of the sentence, i.e., the mean informativeness rating assigned by Turkers in our second MTurk experiment.
\item \textbf{Im:} Implicature of the sentence, i.e., the mean implicature rating assigned by Turkers in our second MTurk experiment.
\item \textbf{Lw:} Length of the sentence in words.
\item \textbf{Lc:} Length of the sentence in characters.
\item \textbf{F:} Formality score of the sentence, as proposed by \newcite{Heylighen99formalityof}.
\item \textbf{I:} Informativeness score of the sentence.
\item \textbf{LD:} Lexical density of the sentence \cite{3324247}.
\end{enumerate}

Among these variables, Heylighen and Dewaele's formality score is given by:

\emph{F = (noun frequency + adjective freq. + preposition freq. + article freq. - pronoun freq. - verb freq. - adverb freq. - interjection freq. + 100)/2}

where the frequencies are taken as percentages with respect to the total number of words in the sentence. The inspiration for this score comes from the fact that nouns, adjectives, prepositions, and articles are found to be \emph{non-deictic} in word correlation studies, whereas pronouns, verbs, adverbs, and interjections are found to be \emph{deictic}.\footnote{Conjunctions are deixis-neutral. We used CRFTagger \cite{citeulike:4893931} to part-of-speech-tag our sentences.} F-score measures formality as the amount of \emph{relative non-deixis} present in a sentence (cf. Section \ref{subsec:formality_background}).

Ure's lexical density takes the form:

\emph{LD = (N$_{lex}$/N) $\times$ 100}

where N$_{lex}$ is the number of \emph{lexical tokens} (nouns, adjectives, verbs, adverbs) in the sentence, and N is the total number of words in the sentence.

The \emph{informativeness score} (\textbf{I}) is a scoring formula we propose in this paper. The idea is as follows. Recall from Section \ref{sec:intro} that \emph{contextuality} -- the opposite of \emph{deep formality} -- is affected by both deixis as well as implicature. Although implicature is very hard to quantify, a measure of ``ambiguity'' in a given piece of text can be formulated by counting how many WordNet senses \cite{Miller:1995:WLD:219717.219748} the words in that text carry on average. The more senses words have, the more ambiguous the text is. The \emph{informativeness score} (\textbf{I}) of a sentence is thus given by the \emph{average number of WordNet senses per word in the sentence}.\footnote{More accurately, it should be called an \emph{ambiguity score}.}

Correlations between the eight variables are given in Table \ref{tab:big_corr}. Note from Table \ref{tab:big_corr} that formality and informativeness are highly correlated in all cases, thereby validating Heylighen and Dewaele's hypothesis that the purpose of formality (\emph{deep formality} in particular) is \emph{more informative communication}. Note, however, that in most cases, there is very little correlation between formality and implicature (small positive/negative values). There are two possible reasons for this: (a) implicature is a poorly-understood phenomenon, and maybe formality and implicature are not as antagonistically related as argued by Heylighen and Dewaele; (b) our implicature annotation by Turkers showed a \emph{central tendency bias} and poor agreement between two MTurk experiments, so maybe the mean implicature ratings we obtained are not truly reflective of the actual amount of implicature present in a sentence. Validating which of these two (or maybe both) is the correct reason, is a part of our future work.

Note further from Table \ref{tab:big_corr} that formality and informativeness are positively correlated (moderate-to-good correlation) with length of the sentence -- in words and characters. This corroborates the earlier finding by \newcite{Lahiri:2011:IJS:1964750.1964792} that as a piece of text gets more formal, it tends to become longer and more intricate. Formality and informativeness also correlate positively (moderate correlation) with Heylighen and Dewaele's F-score, except in the Forum genre. On the other hand, they do not have significant correlations with the informativeness (\textbf{I}) score except the Forum genre. Implicature has a significant, but small negative correlation with F-score in all cases. Lexical density negatively correlates with length of the sentence (\#words and \#characters). Informativeness score correlates positively with length, but negatively with Heylighen and Dewaele's F-score, as expected. Implicature also correlates negatively with F-score in all cases. The two length scores have an almost perfect positive correlation among them, which is unsurprising.

The surprising part, however, is that formality and informativeness (as rated by humans) are not very highly correlated (either positively or negatively) with Heylighen and Dewaele's F-score or our informativeness (\textbf{I}) score. Maybe these two scores are measuring complementary aspects of the phenomenon of formality, and are not individually able to explain all the variations. Automated scoring/prediction of formality by modeling it on top of scores like these (perhaps as features) is our future plan. We would also like to investigate how to predict informativeness, and how to get a better handle on implicature scoring -- both by humans as well as automated.

\subsection{Sentential Make-up of Documents}
\label{subsec:sentential_makeup}

In our final experiment, we investigated how the sentences in a document vary in terms of formality, implicature, and informativeness -- starting from the beginning sentences, then the middle ones, and finally the last ones. We divided the sentences into ten successive bins (\emph{deciles}) based on their position in the document, and measured the mean formality, informativeness, and implicature \emph{per decile}. The results -- averaged across all documents in a particular genre (blog, forums, news, overall) -- are shown in Figure \ref{fig:fii_sent}. Figure \ref{fig:fii_sent} also shows the standard errors for each decile.

Note from Figure \ref{fig:fii_sent} that news sentences are most formal and most informative, followed by blog sentences, followed by forum sentences. In terms of formality and informativeness trends, news sentences start with high formality and informativeness, then gradually diminish in both -- perhaps reflecting the fact that in journalistic writing, first few sentences carry the most information (to catch the readers' attention), and the information/interesting-ness content decreases substantially thereafter. Forum sentences, on the other hand, maintain a low level of formality and informativeness throughout -- with a few small peaks and valleys in-between. For blogs, the trend is first decreasing, then increasing, and then decreasing again -- indicating that the most informative (and formal) sentences in blogs may be in the middle. All three genres taken together, both formality and informativeness show a decreasing trend. There is no clear trend in the implicature rating of sentences -- it is mostly an assortment of peaks and valleys.

\section{Conclusion}
\label{sec:conclusion}

In this paper, we introduced a dataset of 7,032 sentences rated for formality, informativeness, and implicature on a 1-7 scale by human annotators on Amazon Mechanical Turk. To the best of our knowledge, this is the first large-scale annotation effort that ties together all three pragmatic variables at the sentence level. We measured reliability of our annotations by running two independent rounds of annotation on MTurk, and inspecting the correlation among mean ratings between the two rounds. We further examined correlation of our annotations with pilot sentence formality annotations done in a more controlled setting \cite{DBLP:journals/corr/abs-1109-0069}. It was observed that while formality and informativeness can be reliably annotated on a 1-7 scale, implicature poses a much more difficult challenge. We analyzed the distribution of formality, informativeness, and implicature across three genres (news, blogs, and forums), and found significant differences -- both in terms of overall distribution, and also in terms of the documents' sentential make-up. Correlations between the human ratings and five other stylistic variables were carefully examined. 
Our future plans include an automatic sentence-level formality and informativeness predictor, in the same spirit as \cite{conf/acl/Danescu-Niculescu-MizilSJLP13}. We also plan to investigate implicature rating more thoroughly, and figure out a good way to improve reliability in implicature annotation.

The limitations of our study mostly stem from our lack of control on the MTurk experiments. Some of that is intentional, because we really wanted to observe what people think/feel as formal, informative, and implicative. However, previous studies have employed measures like background questionnaires, linguistic attentiveness surveys, and z-scoring to weed out/smooth difficulties \cite{conf/acl/Danescu-Niculescu-MizilSJLP13}. While these are indeed promising research directions to try, we opine that even without such stringent measures, we were able to obtain quite good annotations -- except implicature, where the earlier approach of \newcite{degen2015} may truly be very helpful.

\section*{Acknowledgments}
\label{sec:acknowledgments}

We gratefully acknowledge Rada Mihalcea for her support; MTurk annotators for their annotations; Eduard Hovy and Julian Brooke for valuable discussions; Haiying Li and Nina Dethlefs for inspiration and dataset; Francis Heylighen and Jean-Marc Dewaele for their kindness and brilliant ideas, including the I-score; and lastly but most importantly, Xiaofei Lu for his continuous encouragement, warm intellectual companionship, excellent advice and camaraderie, sound ideas in the early stages of the study, and great help with thought processing. This work would not have been possible without you. All results, discussions, and comments contained herein are the sole responsibility of the author, and in no way associated with any of the above-mentioned people. The errors and omissions, if any, should be addressed to the author, and will be thankfully received.

\bibliographystyle{acl2012}
\bibliography{formality_bibliography}

\end{document}